\documentclass{article}

      \usepackage[nonatbib, preprint]{neurips_2025}

\usepackage[utf8]{inputenc} % allow utf-8 input
\usepackage[T1]{fontenc}    % use 8-bit T1 fonts
\usepackage{hyperref}       % hyperlinks
\usepackage{url}            % simple URL typesetting
\usepackage{booktabs}       % professional-quality tables
\usepackage{amsfonts}       % blackboard math symbols
\usepackage{nicefrac}       % compact symbols for 1/2, etc.
\usepackage{microtype}      % microtypography
\usepackage{xcolor}         % colors
\usepackage{amsmath}
\usepackage{graphicx} %插入图片的宏包
\usepackage{float} %设置图片浮动位置的宏包
\usepackage{subfigure} %插入多图时用子图显示的宏包
\usepackage{multirow}
\usepackage{array}
\usepackage{colortbl}
\usepackage{mathrsfs}
\usepackage{amsthm}
\usepackage{amssymb}
\usepackage{graphicx}

\title{RFTF: Reinforcement Fine-tuning for Embodied Agents with Temporal Feedback}

\author{%
  Junyang Shu\thanks{These authors contributed equally to this work.}~~ \quad
  Zhiwei Lin$^{*}$ \quad 
  Yongtao Wang\thanks{Corresponding author} \\
  Wangxuan Institute of Computer Technology, Peking University, China\\
  \texttt{\{zwlin, wyt\}@pku.edu.cn} \\
}

\begin{document}

\maketitle

\begin{figure}[htpb]
    \centering
    \includegraphics[width=1\linewidth]{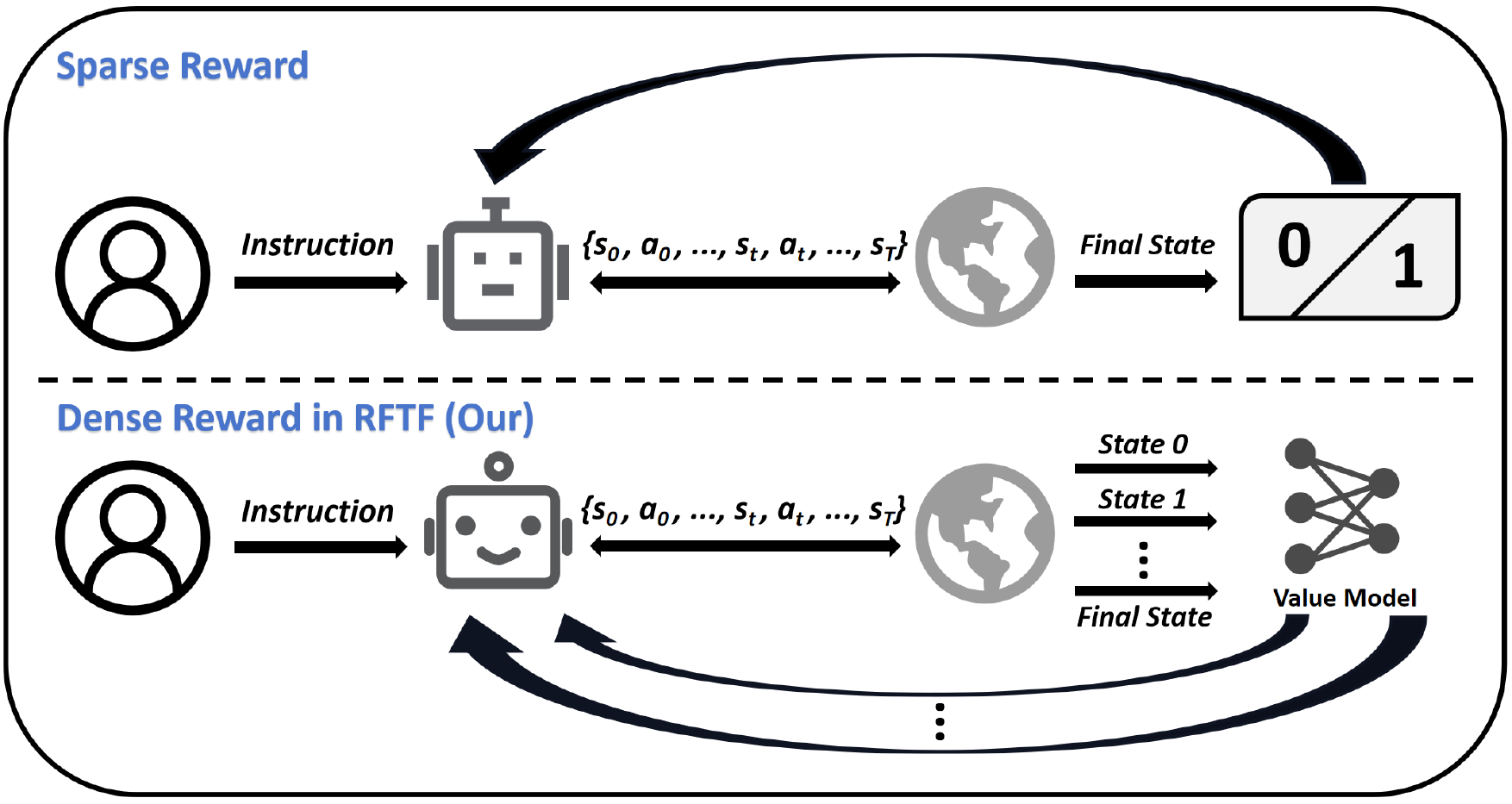}
    \caption{\textbf{Comparison between sparse reward and dense reward.}
    In typical reinforcement fine-tuning methods for embodied agents, only sparse, outcome-based rewards are provided, which can confuse embodied agents when encountering partially correct or incorrect episodes. In contrast, RFTF leverages a value model trained with temporal information to predict the value of each state within an episode, providing embodied agents with higher-granularity dense rewards.}
    \label{sparse_vs_dense}
\end{figure}

\begin{abstract}
Vision-Language-Action (VLA) models have demonstrated significant potential in the field of embodied intelligence, enabling agents to follow human instructions to complete complex tasks in physical environments. 
Existing embodied agents are often trained through behavior cloning, which requires expensive data and computational resources and is constrained by human demonstrations.
To address this issue, many researchers explore the application of reinforcement fine-tuning to embodied agents. 
However, typical reinforcement fine-tuning methods for embodied agents usually rely on sparse, outcome-based rewards, which struggle to provide fine-grained feedback for specific actions within an episode, thus limiting the model’s manipulation capabilities and generalization performance. 
In this paper, we propose RFTF, a novel reinforcement fine-tuning method that leverages a value model to generate dense rewards in embodied scenarios. 
Specifically, our value model is trained using temporal information, eliminating the need for costly robot action labels. 
% 这里需要再具体说techniques是啥 DONE
In addition, RFTF incorporates a range of techniques, such as GAE and sample balance to enhance the effectiveness of the fine-tuning process. 
% 这部分可以挪到experimental results前面 DONE
By addressing the sparse reward problem in reinforcement fine-tuning, our method significantly improves the performance of embodied agents, delivering superior generalization and adaptation capabilities across diverse embodied tasks.
Experimental results show that embodied agents fine-tuned with RFTF achieve new state-of-the-art performance on the challenging CALVIN ABC-D with an average success length of $4.296$.
% 下面的$4.301$指标也改一下，具体一点 DONE
Moreover, RFTF enables rapid adaptation to new environments. After fine-tuning in the D environment of CALVIN for a few episodes, RFTF achieved an average success length of $4.301$ in this new environment.

\end{abstract}

\section{Introduction}
\label{Introduction}

% 引用cite之前要加 空格或~ DONE
The field of embodied intelligence has made remarkable progress in recent years~\cite{liu2024aligning}. 
Vision-Language-Action (VLA) models, by integrating visual perception, language understanding, and action execution, empower agents to perform a variety of tasks in the physical world following human instructions~\cite{ma2024survey}. 
% RL算是一种supervised learning吗？ data最好改成labeled data或者其他需要带额外标注的 DONE
% However, training embodied agents directly through supervised learning or behavior cloning requires vast and costly data and computational resources~\cite{bu2025agibot, o2024open}, and is limited by the scope of human demonstrations. 
However, training embodied agents directly through behavior cloning requires vast and costly labeled data and computational resources~\cite{bu2025agibot, o2024open}, and is limited by the scope of human demonstrations. 
% 我们的方法有改善reasoning吗？没有的话就不要提 DONE
% As a result, embodied agents often exhibit suboptimal generalization and reasoning capabilities in practical applications.
As a result, embodied agents often exhibit suboptimal manipulation and generalization capabilities in practical applications.

To enhance the generalization and reasoning abilities of embodied agents while reducing reliance on extensive data and computational resources, increasing research efforts have focused on reinforcement fine-tuning for embodied agents~\cite{chen2025conrft, ren2024diffusion, hu2024flare, guo2025improving}. 
% RL缩写第一次出现需要标注 DONE
% RL trains models through trial-and-error, allowing them to learn from past experiences without depending on human-provided data. 
Reinforcement learning (RL) trains models through trial-and-error, allowing them to learn from past experiences without depending on human-provided data. 
In addition, RL allows models to adapt to new environments by trying out and updating the parameters without human intervention.
However, a significant challenge in reinforcement fine-tuning for embodied agents is the reliance on sparse, outcome-based rewards. 
Such reward mechanisms fail to capture the nuanced correctness of individual actions within an episode. 
For instance, when there is partial correctness or incorrectness, it can lead to erroneous encouragement or suppression of certain actions, ultimately compromising the efficiency and stability of reinforcement fine-tuning~\cite{lyu2025exploring}.

% In this paper, we propose a two-stage approach to achieve dense-reward reinforcement fine-tuning for embodied agents. 
In this paper, we propose RFTF, a reinforcement fine-tuning method with dense-reward from temporal feedback for embodied agents. 
RFTF contains two stages. 
Specifically, in the first stage, we present a value model tailored for embodied scenarios. 
This value model is trained using temporal information to predict the value of the current state based on the input state and human instruction.
% , notably without requiring expensive robot action labels. 
%
In the second stage, we integrate this value model into an RL fine-tuning framework for embodied agents based on Proximal Policy Optimization (PPO)~\cite{schulman2017proximal}. 
By combining reward shaping and generalized advantage estimation (GAE)~\cite{schulman2015high}, we ensure an efficient reinforcement fine-tuning process. 
Notably, the entire training process for the value model and embodied agent does not require robot action labels, showing the proposed method's potential for efficient data utilization.
% Since no robot action labels are involved, our approach not only provides dense rewards to address the need for fine-grained feedback but also maintains data efficiency, making it an effective solution for embodied tasks.

The contributions of this paper are as follows:
\begin{itemize}
\item We introduce a dense-reward reinforcement fine-tuning method for embodied agents without any robot action labels from humans.

\item We present a value model trained with temporal information to generate dense rewards and combine reward shaping and GAE strategy to facilitate the RL fine-tuning process.

\item Experimental results on the CALVIN benchmark~\cite{mees2022calvin} show that the proposed method achieves new state-of-the-art performance under the ABC-D setting. 
Moreover, RFTF exhibits superior adaptation capabilities in a new environment.
\end{itemize}

\section{Related work}
\label{Related work}
\subsection{Foundation models for embodied agents}
\label{Foundation models for embodied agents}
In recent years, large language models (LLMs)~\cite{achiam2023gpt, yang2024qwen2, touvron2023llama, guo2025deepseek} and vision-language models (VLMs)~\cite{karamcheti2024prismatic, liu2023visual, chen2024internvl, team2024gemini, anthropic2024claude} trained on web-scale data have not only demonstrated the ability to engage in human-like dialogue but also exhibited remarkable reasoning capabilities and understanding of the physical world~\cite{huang2022inner, szot2023large, wu2023embodied, li2024manipllm, mazzaglia2024genrl}. 
% i.e.;e.g.这种需要加斜体 DONE
Leveraging these strengths, many researchers explore how to apply foundation models to interact with the physical world, \textit{i.e.}, for embodied intelligence.

Broadly, there are two approaches to achieve physical environment interaction with foundation models: high-level planning and low-level manipulation. 
For high-level planning, LLMs or VLMs are used for planning to capitalize on their robust comprehension and reasoning abilities to translate complex human instructions into simpler robot skills~\cite{brohan2023can, hu2023look, ji2025robobrain, mao2024robomatrix, wu2023embodiedplanning}, \textit{e.g.}, move to, grasp, and so on. 
However, directly utilizing LLMs or VLMs does not yield control signals for embodiments and requires additional models to complete simpler robot skills.  
For low-level manipulation, the output of the pretrained VLMs is changed to action and construct VLAs, enabling direct interaction with the physical environment. 
%
% Given that VLMs can achieve unified encoding of visual and linguistic information, numerous studies have proposed modifying the output part of pretrained VLMs to construct VLAs, enabling direct interaction with the physical environment. 
%
Specifically, RT-2~\cite{zitkovich2023rt} fine-tunes PaLI-X~\cite{chen2023pali} on both vision-language data and robot demonstrations to build the robot policy. 
OpenVLA~\cite{kim2024openvla} fine-tunes Prismatic on Open X-Embodiment (OXE)~\cite{o2024openx} dataset, which contains 22 different robotic embodiments from 21 different institutions. 
$\pi_0$~\cite{black2024pi_0} proposes a novel flow matching architecture built on top of the PaliGemma VLM~\cite{beyer2024paligemma} to inherit Internet-scale semantic knowledge. 
It is worth noting that not all VLAs are derived from pre-trained VLMs. 
RDT~\cite{liu2024rdt} builds on diffusion models to perform complex bimanual manipulation. 
Seer~\cite{tian2024predictive} generates actions through an inverse dynamics model that is conditioned on the robot's anticipated visual states.

However, these models are trained with human action labels in a behavior cloning way.
In this paper, we train VLAs through reinforcement fine-tuning to improve their generalization and help them adapt to novel environments with any action labels.
% we improve the generalization capability of VLAs through reinforcement fine-tuning, which can also help them adapt to novel environments.

% 讲一下我们跟之前方法的不同 DONE（一句话比较短，加在第二段末尾了）
% In this paper, we xxx

\subsection{Reinforcement fine-tuning for large models}
\label{Reinforcement fine-tuning for large models}
Reinforcement learning is a technique that enables learning from a model's past experiences. Unlike supervised learning, RL does not require large amounts of manually annotated data, nor is it constrained by expert demonstrations. 
Recently, using RL to fine-tune pretrained large language models has become a trend, enhancing models' reasoning capabilities or aligning them with human preferences~\cite{ouyang2022training, guo2025deepseek, shao2024deepseekmath, trung2024reft, zhai2024fine}. 
However, reinforcement fine-tuning in the field of embodied intelligence differs from that in LLMs, as it necessitates extensive interaction with the environment to collect data. 
FLaRe~\cite{hu2024flare} is a large scale reinforcement fine-tuning framework that introduces a series of design choices that help stabilize the RL training process.
iRe-VLA~\cite{guo2025improving} iterates between reinforcement learning and supervised learning to address the instability issues of reinforcement learning in large-scale VLAs. 
DPPO~\cite{ren2024diffusion} improves diffusion-based policies by leveraging the sequential nature of the diffusion denoising process and fine-tuning the entire chain of diffusion MDPs.

% 讲一下说我们将dense reward引入了vla DONE
Nonetheless, due to the high precision requirements for output actions, directly applying sparse rewards in the reinforcement fine-tuning of embodied agents often yields suboptimal results and may even lead to performance degradation.
In contrast, we incorporate dense rewards into the reinforcement fine-tuning of embodied agents through a value model trained with temporal information. 
%
% Specifically, our value model predicts the state value at each time step, thereby providing more fine-grained guidance to embodied agents than sparse rewards.

% 这部分可以合到method里面，改成sec: notation and Preliminary DONE

\section{Method}
\label{Method}
Our objective is to utilize a value model to provide dense rewards for reinforcement fine-tuning of embodied agents, thereby enhancing their generalization and adaptation capabilities. 
In Section \ref{Notation and preliminary}, we begin by introducing the notation and preliminary used in our approach. In Section \ref{Value model}, we introduce the structure of the value model and the methodology for its training. 
In Section \ref{Fine-tuning pipeline}, we describe how the value model is applied to the reinforcement fine-tuning process for embodied agents.

\subsection{Notation and preliminary}
\label{Notation and preliminary}
Due to challenges such as limited camera coverage and occlusions between objects, we model each robot task as a Partially Observable Markov Decision Process (POMDP), defined by the tuple $(\mathcal{S}, \mathcal{A}, \mathcal{P}, \mathcal{R}, \mathcal{O}, \mathcal{L}, \gamma)$. Here, $\mathcal{S}$ and $\mathcal{A}$ represent the state space and action space, respectively. 
$\mathcal{P}: \mathcal{S} \times \mathcal{A} \times \mathcal{S} \rightarrow [0, 1]$ denotes the state transition probability function.
$\mathcal{R}: \mathcal{S} \times \mathcal{A} \times \mathcal{L} \times \mathcal{S} \rightarrow \mathbb{R}$ is the reward function. 
$\mathcal{O}$ is the observation space, specifically consisting of RGB images from different cameras. 
$\mathcal{L}$ is the set of human instructions guiding the robot to complete tasks. 
$\gamma \in [0, 1]$ is the discount factor, controlling the balance between short-term and long-term focus.

We denote the current robot policy as $\pi_\theta$, \textit{i.e.}, a Vision-Language-Action model parameterized by $\theta$. The policy $\pi_\theta$ takes the state (\textit{e.g.}, observation and robot state) as input and outputs the corresponding action. 
The overall optimization objective is to maximize the expected return of $\pi_\theta$ with the discount factor $\gamma$, \textit{i.e.}, $J(\theta) = \mathbb{E}_{(s_t, a_t)\sim P}\sum_{t}\gamma^t R(s_t, a_t)$.

\subsection{Value model}
\label{Value model}
\begin{figure}[t]
    \centering
    \includegraphics[width=1\linewidth]{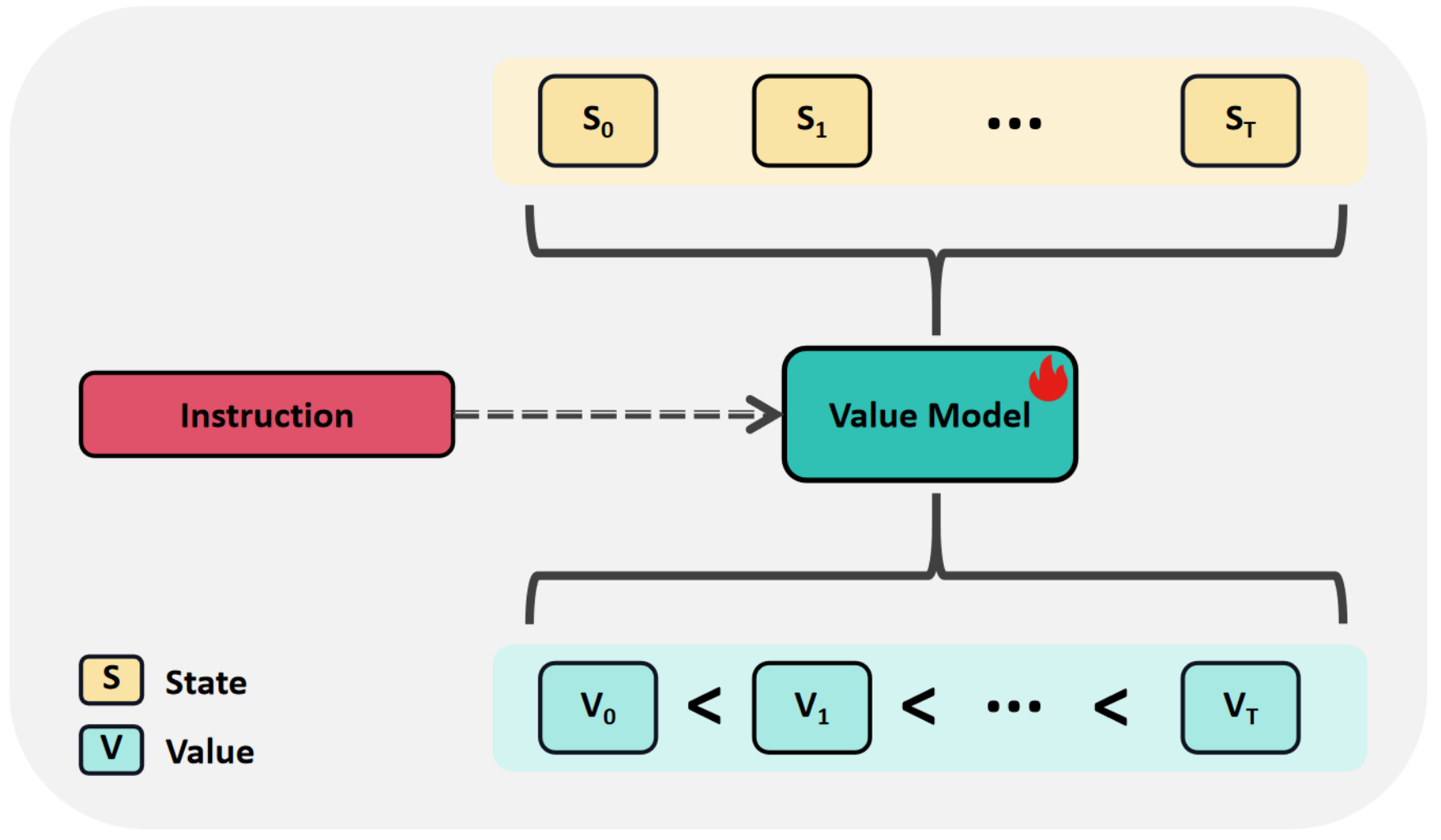}
    \caption{\textbf{Training procedure of the value model}. We assume that during an episode of a human-demonstrated successful embodied task, the state value increases monotonically over time.}
    \label{value_model_training}
\end{figure}
Embodied agents take the human instructions and state at each time step as input, outputting low-level robot actions to interact with the environment and complete tasks. 
An embodied episode may consist of hundreds or thousands of actions. Current reinforcement fine-tuning methods applied to embodied agents, such as FLaRe~\cite{hu2024flare} and DPPO~\cite{ren2024diffusion}, predominantly rely on sparse, outcome-based rewards, which are inadequate when the decisions of embodied agents exhibit local correctness or errors.

To automatically label the correctness of each action in an episode and provide dense reward for training, we employ a value model parameterized by $\phi$ to predict the value of the state at each time step. 
Specifically, the value model takes the state and human instruction as input and outputs the value of the current state, \textit{i.e.}, $v_t = V_{\phi}(s_t, l)$.
Notably, both the inference and training of the value model do not require expensive robot action labels, resulting in low data dependency.

% However, there are currently no direct value labels available for training the value model. 
However, there are no explicit value labels available for training the value model. 
Inspired by RLHF~\cite{ouyang2022training}, we collect data pairs and train the value model with contrastive learning. 
Specifically, as shown in Figure~\ref{value_model_training}, $(s_t, s_{t+1}, ..., s_{t+n-1}~|~l)$ is an expert-demonstrated trajectory without action labels. 
As the state progresses toward task completion, we assume that the state value increases with each time step, \textit{i.e.}, $v_t < v_{t + 1} < ... < v_{t+n-1}$.
Following RLHF, we adopt the contrastive loss function as the optimization objective:
\begin{equation}\label{1}
    loss(\phi) = - \frac{1}{C_{n}^{2}}\mathbb{E}_{(s_t, a_t)\sim P}[log(\sigma(V_{\phi}(s_{t+\Delta t}, l) - V_{\phi}(s_t, l))],
\end{equation}
where $C_{n}^{2}$ is the number of combinations, $\sigma$ is the sigmoid function and $\Delta t$ is a positive integer belonging to $[1, n-t)$. 
This implies that we aim for the value at later time steps to be as large as possible compared to earlier time steps. 
% C_n^2也要解释一下 DONE

The architecture of our value model is based on the VLA model, with only the action tokens of the VLA model replaced by value tokens. Given that the VLA model is already capable of processing mixed inputs of states and human instructions, we initialize the training of the value model using the weights of the VLA model.

\subsection{RL fine-tuning pipeline}
\label{Fine-tuning pipeline}
\begin{figure}[t]
    \centering
    \includegraphics[width=1\linewidth]{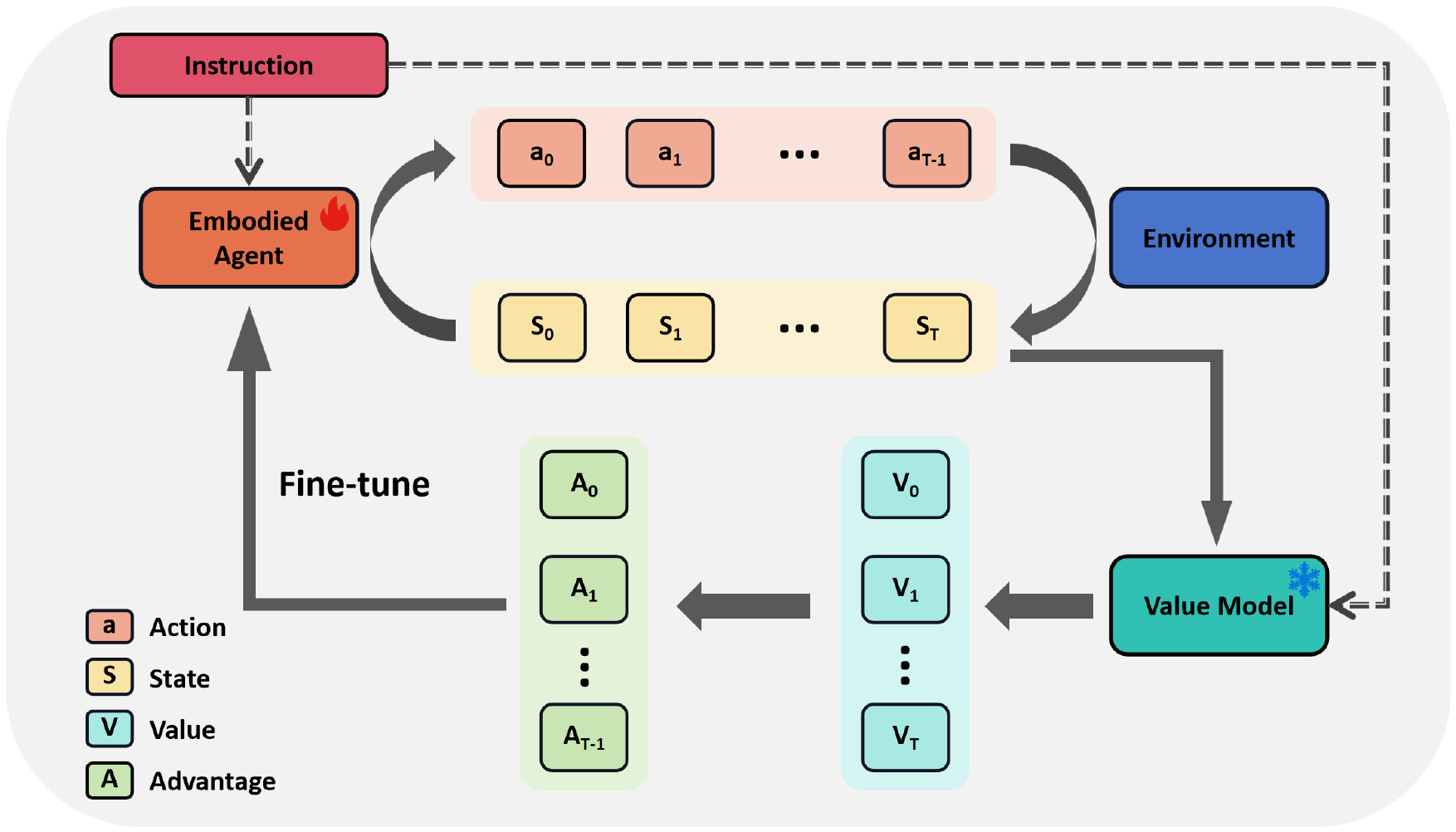}
    \caption{
    % We propose RFTF, a reinforcement fine-tuning framework for embodied agents. 
    \textbf{Illustration of RL fine-tuning pipeline.}
    RFTF utilizes a value model trained with temporal information to predict the value of each state in episodes of interaction between the embodied agent and the environment, thereby providing guidance for each action in episodes to fine-tune the embodied agent.}
    \label{pipeline}
\end{figure}
In this stage, as shown in Figure \ref{pipeline}, we utilize the trained value model to guide the RL fine-tuning process, aiming to enhance the performance of embodied agents. To achieve effective fine-tuning, we adopt Proximal Policy Optimization (PPO)~\cite{schulman2017proximal} as our reinforcement learning algorithm framework.

Specifically, after obtaining the raw value of each state in an episode from the trained value model, we normalize all state values within the episode, as the output range of the value model varies significantly across different tasks. To make progress toward the final goal, we employ a reward shaping term~\cite{ng1999policy} as the reward function:
\begin{equation}\label{2}
    R_t = 
        \begin{cases}
            \gamma V(s_{t+1}, l) - V(s_t, l) & \text{if}~t~\text{is not the end step} \\
            0 & \text{otherwise}
        \end{cases}
\end{equation}

Notably, if we directly use the discounted sum of rewards with the discount factor $\gamma$ as the advantage function in PPO, the simplified form of the advantage function will contain only the current and last states, thereby neglecting the influence of intermediate states. To address this, we adopt Generalized Advantage Estimation (GAE)~\cite{schulman2015high}, introducing an additional hyperparameter $\lambda$ to incorporate the state values at all intermediate time steps into the advantage function. 
Additionally, to leverage information about task completion, instead of solely adding +1 or -1 to the reward at the final time step, we incorporate feedback on task success or failure into the advantage function at each time step. This ensures that early decisions in long episodes receive relevant feedback. 
Furthermore, we introduce a balancing coefficient to address the imbalance between successful and failed samples. The final form of our advantage function is as follows:
\begin{equation}\label{3}
    A_t = \eta\left[I(success) + \sum_{n=t}^{T}(\gamma \lambda)^{n - t} R_t\right],
\end{equation}
where $\eta$ is the coefficient for balancing positive and negative samples, set to 0.25 when the task succeeds and 1 when the task fails. $\lambda$ is the hyperparameter in GAE. $I$ is an indicator function that returns +1 for task success and -1 for task failure.

Previous works~\cite{hu2024flare, guo2025improving, chen2025conrft} have highlighted that, due to the high precision requirements for embodied agent outputs, directly applying reinforcement fine-tuning to embodied agents often results in performance degradation. To mitigate this, inspired by~\cite{shao2024deepseekmath}, we incorporate both a surrogate objective clipping term and adaptive KL divergence into the optimization objective of reinforcement fine-tuning. Specifically, for an embodied agent $\pi$ parameterized by $\theta$, the loss function for reinforcement fine-tuning is as follows:
\begin{align}
    loss(\theta) =  -\mathbb{E}_{(s_t, a_t)\sim P} 
    \Big \{ min\Big [\frac{\pi_\theta (a_t | s_t)}{\pi_{\theta_{old}}(a_t | s_t)}A_t, clip\Big (\frac{\pi_\theta (a_t | s_t)}{\pi_{\theta_{old}}(a_t | s_t)}, 1 - \epsilon, 1 + \epsilon \Big )A_t \Big] \notag \\ -
    \beta D_{KL} \left [\pi_\theta (a_t | s_t) || \pi_{\theta_{ref}}(a_t | s_t) \right ] \Big \} ,
\end{align}
where $\epsilon$ and $\beta$ are hyper-parameters that prevent excessive divergence between the new and old policies.

\section{Experiments}
\label{Experiments}
% In this section, we conducted extensive experiments on the challenging CALVIN benchmark~\cite{mees2022calvin}. Through these experiments, we aim to address the following three questions:
% \begin{itemize}
% \item Can the value model accurately assess the value of a state?
% \item Can RFTF enable embodied agents to adapt to new environments or improve their generalization?
% \item Can the dense rewards provided by RFTF outperform sparse rewards in reinforcement fine-tuning?
% \end{itemize}

\subsection{Experimental setup}
\label{Experimental setup}
\subsubsection{Benchmark}
\label{Benchmark}
\begin{figure}[t]
    \centering
    \subfigure[Env A]{
    \includegraphics[width=0.23\linewidth]{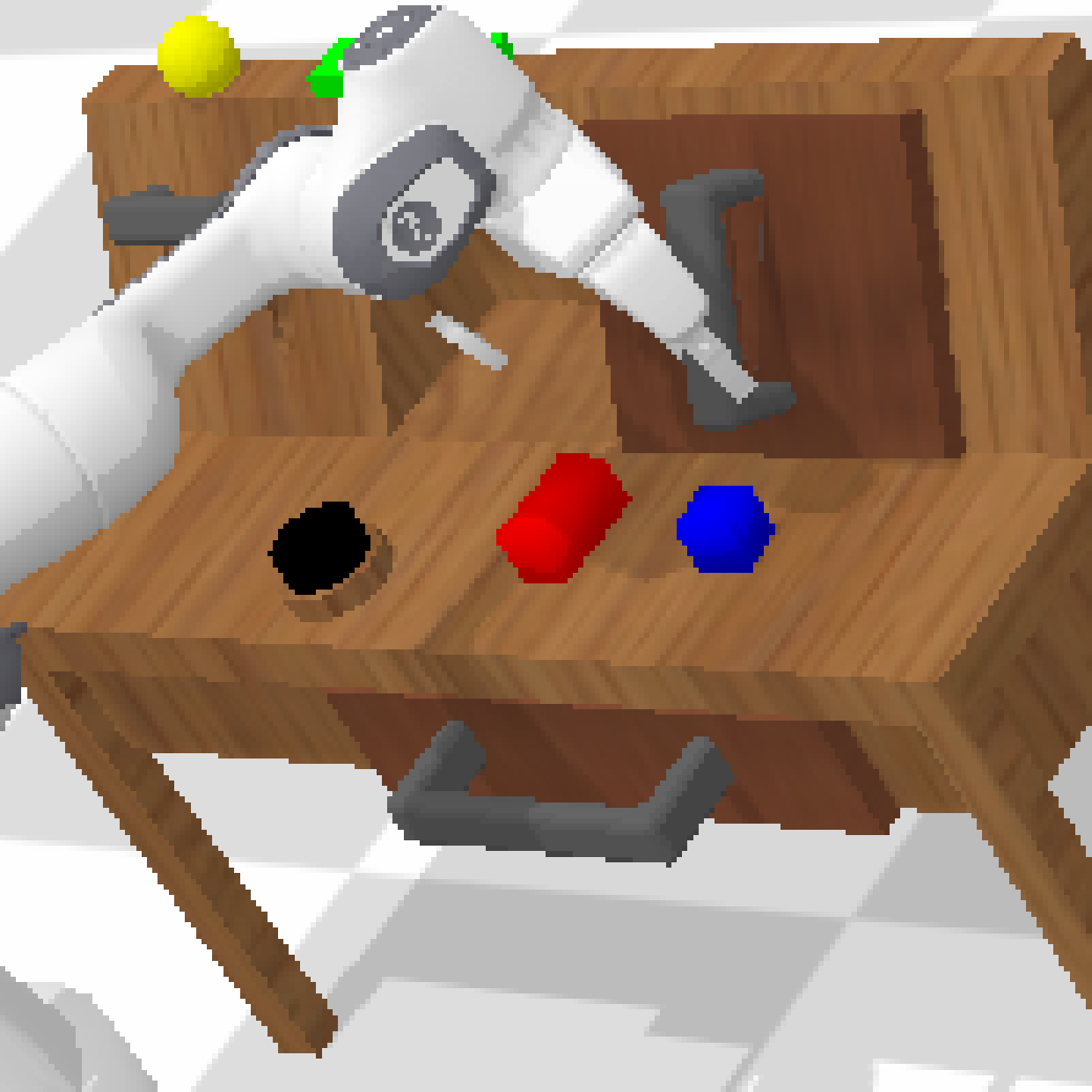}
    }
    \subfigure[Env B]{
    \includegraphics[width=0.23\linewidth]{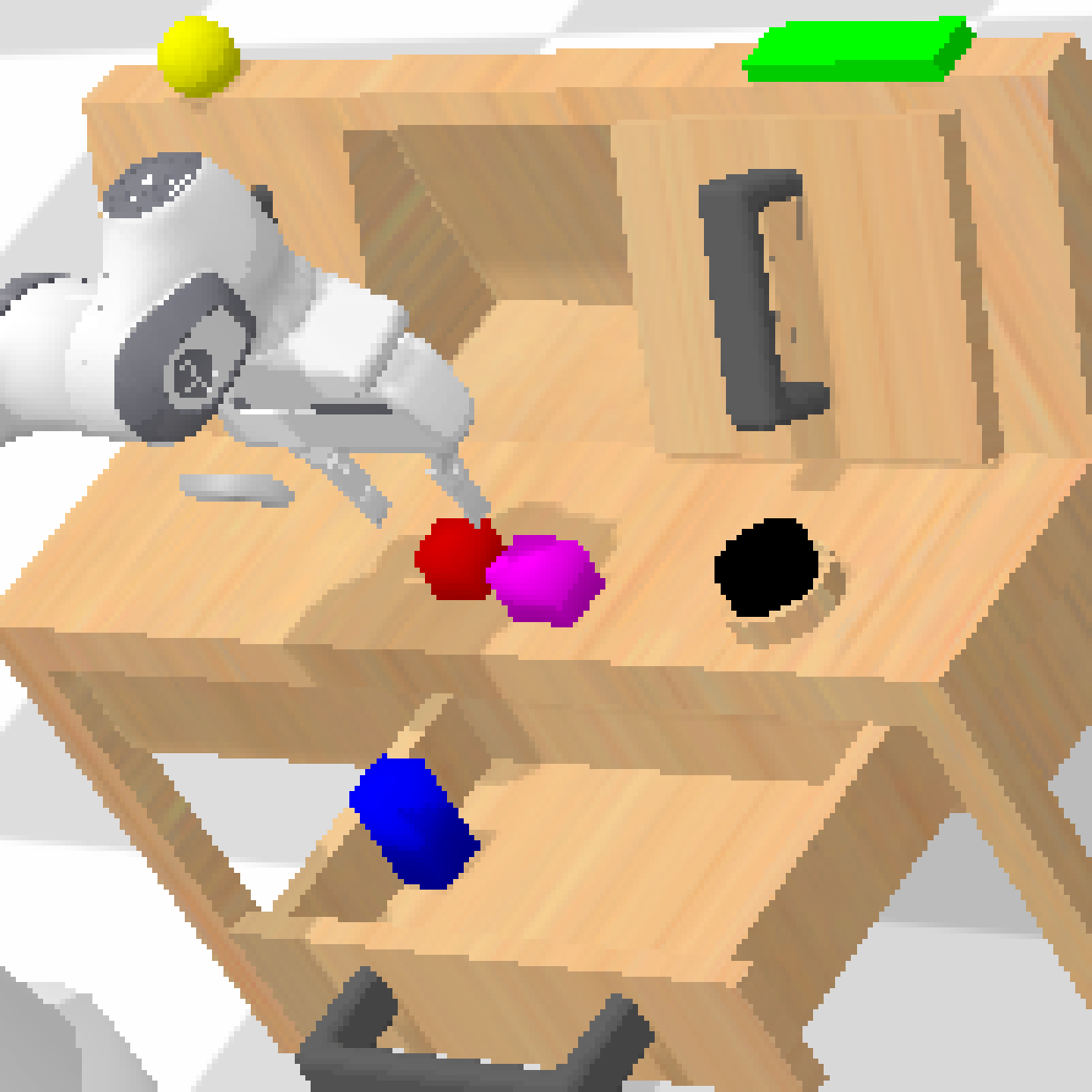}
    }
    \subfigure[Env C]{
    \includegraphics[width=0.23\linewidth]{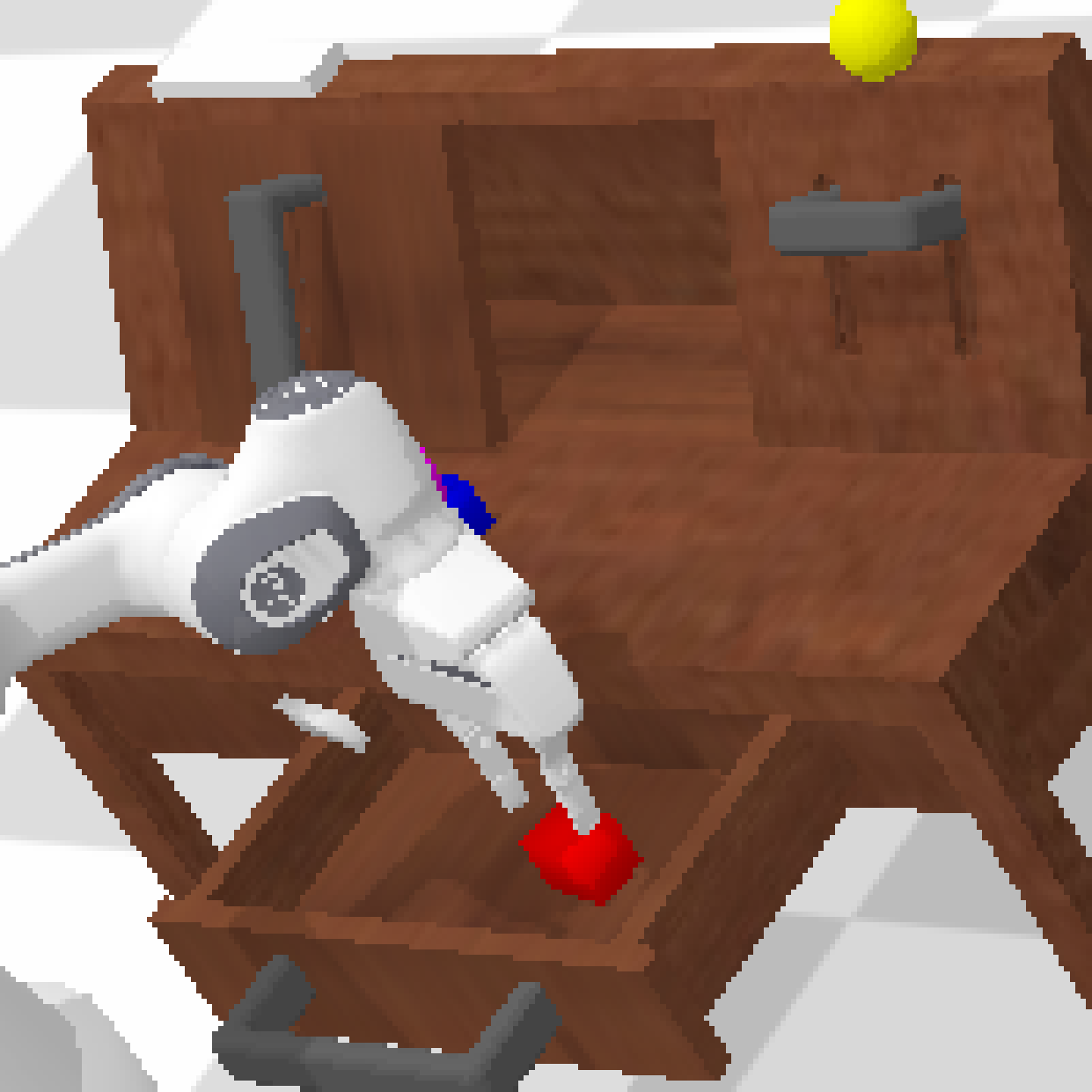}
    }
    \subfigure[Env D]{
    \includegraphics[width=0.23\linewidth]{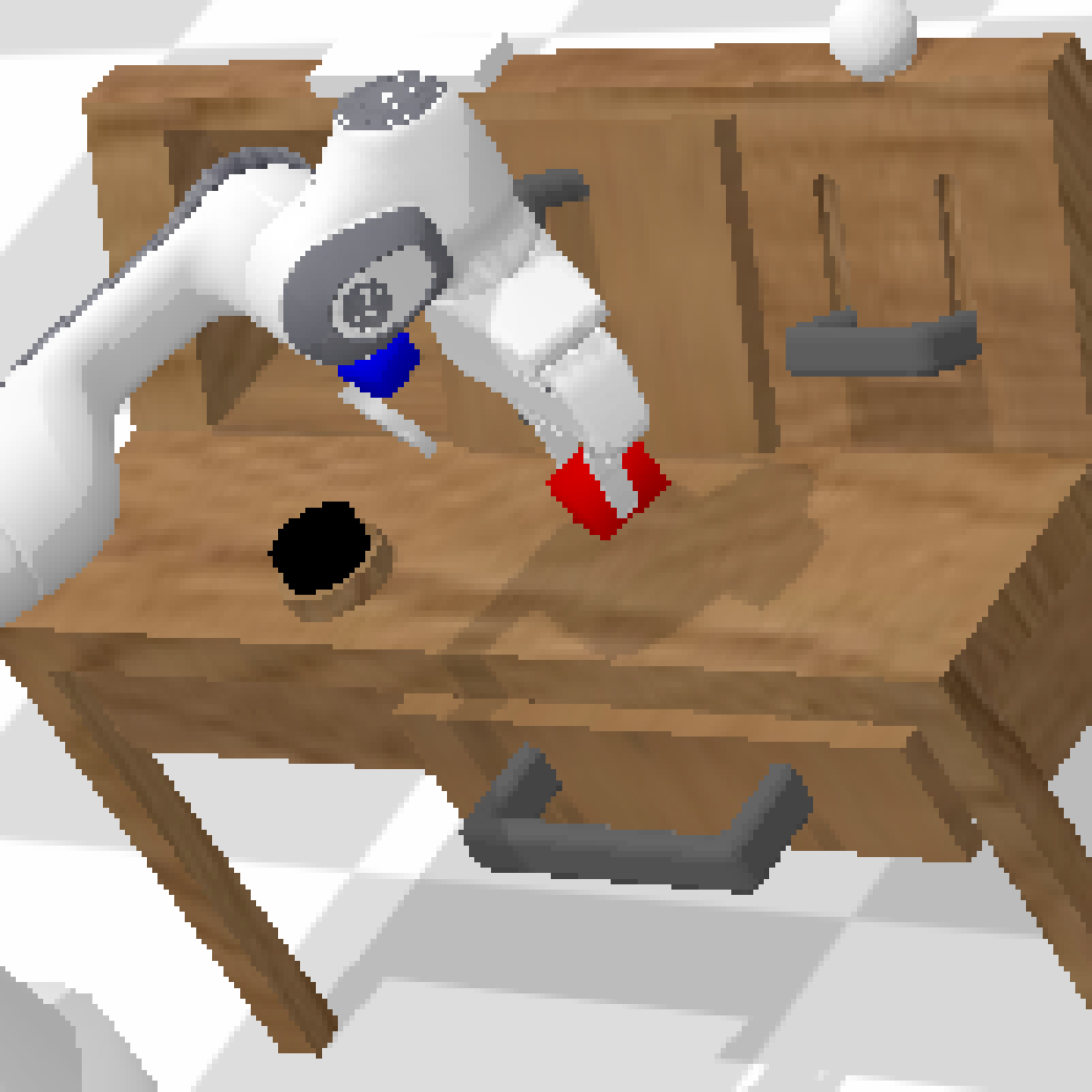}
    }
    \caption{
    \textbf{Visualization of the CALVIN benchmark}.
    The CALVIN benchmark includes four distinct environments, differing in the positions of the LED, light bulb, slider, drawer, switch, and button, as well as the material of the table.}
    \label{calvin_env}
\end{figure}
We conducted experiments on the challenging CALVIN benchmark. CALVIN focuses on long-horizon robotic tasks conditioned on language, encompassing 34 tasks across four distinct simulated environments (Env A, B, C, and D). As shown in Figure \ref{calvin_env}, each environment features a Franka Emika Panda robot equipped with a parallel-jaw gripper and a table for performing various manipulation tasks. Leveraging the four diverse environments of the CALVIN benchmark, we can conduct both generalization and adaptation experiments to validate the effectiveness of our approach.
\subsubsection{Baselines}
\label{Baselines}
% We selected top-performing models from the CALVIN leaderboard as baselines. The 3D Diffuser Actor~\cite{ke20243d} captures 3D features for embodied tasks; CLOVER~\cite{bu2024closed} introduces a feedback mechanism to enhance robotic manipulation capabilities; the Diffusion Transformer Policy~\cite{hou2024diffusion} directly achieves action denoising generation through a large-scale transformer; GR-MG~\cite{li2025gr} employs a diffusion model to predict future observations, guiding action generation; and Seer~\cite{tian2024predictive}, a predictive inverse dynamics model, simultaneously predicts future states and outputs actions, optimizing the final output actions through intermediate features of future states.
%

In RFTF, we fine-tune the top two models on the CALVIN ABC-D benchmark under two distinct settings. In the generalization setting, we fine-tune the models on the CALVIN ABC environments, which the models have already encountered, to evaluate whether RFTF enhances the models' generalization performance. In the adaptation setting, we fine-tune the models on the CALVIN D environment, which the models have not previously seen, to assess whether RFTF helps the models adapt to new environments.
\subsubsection{Metrics}
\begin{table}[t]
    \centering
    % \caption{\textbf{CALVIN ABC-D results.} Generalization refers to models that have been fine-tuned in CALVIN's A, B, and C environments, while Adaptation refers to models that have been fine-tuned in CALVIN's D environment.}
    \caption{\textbf{CALVIN ABC-D results.} 
    We present the average success rates of top-3 checkpoints computed over 1000 rollouts for each task and the average number of completed tasks to solve 5 instructions consecutively (Avg. Len.). 
    The model been fine-tuned is specified in parentheses after RFTF.}
    \label{generalization_results}
    \scalebox{1}{
        \begin{tabular}{c|ccccc|c}
            \toprule
            \multirow{2}{*}{Method} & \multicolumn{6}{c}{Task completed in a row}  \\
            \cmidrule{2-7}
             & L1 & L2 & L3 & L4 & L5 & Avg. Len. $\uparrow$\\
            \midrule
            3D Diffuser Actor~\cite{ke20243d} & 92.2 & 78.7 & 63.9 & 51.2 & 41.2 & 3.272 \\
            CLOVER~\cite{bu2024closed} & 96.0 & 83.5 & 70.8 & 57.5 & 45.4 & 3.532 \\
            Diffusion Transformer Policy~\cite{hou2024diffusion} & 94.5 & 82.5 & 72.8 & 61.3 & 50.0 & 3.611 \\
            Seer~\cite{tian2024predictive} & 94.4 & 87.2 & 79.9 & 72.2 & 64.3 & 3.980 \\
            GR-MG~\cite{li2025gr} & 96.8 & 89.3 & 81.5 & 72.7 & 64.4 & 4.047 \\
            \textbf{RFTF(GR-MG)} & \textbf{96.9} & \textbf{88.8} & \textbf{82.1} & \textbf{74.9} & \textbf{65.4} & \textbf{4.081} \\
            Seer-Large~\cite{tian2024predictive} & 96.3 & 91.6 & 86.1 & 80.3 & 74.0 & 4.283 \\
            \textbf{RFTF(Seer-Large)} & \textbf{96.4} & \textbf{91.7} & \textbf{86.7} & \textbf{80.7} & \textbf{74.1} & \textbf{4.296} \\
            \bottomrule
        \end{tabular}
    }
\end{table}
\label{Metrics}
The CALVIN benchmark evaluation requires embodied agents to execute 1000 sequences in the D simulated environment, with each sequence comprising 5 tasks. The embodied agent performs these 5 tasks sequentially, exiting the sequence upon failing any task. 
%
% Ln represents the proportion of completing n out of the 5 tasks. 
Ln denotes the proportion of n tasks completed out of 5.
We use the average number of tasks completed per sequence as the evaluation metric.

\subsubsection{Implementation details}
\label{sec Implementation details}
% We conduct all experiments using four NVIDIA A40 GPUs.

For the value model, we train it with a batch size of 4$\times$8 and a learning rate of 1e-5. As described in the Section \ref{Value model}, the structure of the value model involves replacing the original VLA's action tokens and action decoder with value tokens and a value decoder.

For the RL fine-tuning, we apply the same implementation details across both the generalization and adaptation settings to ensure the method's universality. 
% %
% We first discretize the model’s output range into 1000 bins to obtain the probability term in the PPO optimization objective. 
%
For the VLA model, we first discretize the model’s output with 1000 bins to obtain the probability term in the PPO optimization objective. 
To enhance training stability, we freeze the model’s encoders and transformer backbone, and only update the action head. 
During RL fine-tuning, we train the model with a learning rate of 1e-7, covering roughly 1000 episodes.
The RL fine-tuning process is done with four NVIDIA A40 GPUs within 10 hours for Seer-Large and 14 hours for GR-MG.
To prevent overfitting to task instructions, we deliberately used different instructions during the fine-tuning phase than those used during the testing phase.

\subsection{Main results}
\label{Main results}

% Table 1可以只放 RFTF(Seer-Large, Generalization)，名字直接改成RFTF； DONE
% generalization和adaptation是不是重新弄个表格单独讲设置和故事好一些？ DONE
% 
We validate the effectiveness of our algorithm from two perspectives: first, RFTF enhances the generalization capabilities of embodied agents; second, RFTF facilitates adaptation to new environments. Notably, the baseline models used for fine-tuning were trained solely on demonstration data from CALVIN’s A, B, and C environments, without any data from the D environment. To ensure the reliability of the results, each experiment was evaluated using three different seeds, with the mean value reported as the final result.

\subsubsection{Generalization}
\label{Generalization}

% In experiments aimed at improving the generalization ability of embodied agents, we fine-tune the embodied agents in the original training environment. 
In this experiment, we show the better generalization ability of embodied agents fine-tuned with RFTF.
Specifically, RFTF fine-tunes models on CALVIN’s A, B, and C environments and tests them in the D environment. As shown in Table \ref{generalization_results}, GR-MG fine-tuned by RFTF achieved a score of $4.081$, surpassing the baseline of $4.043$; Seer-Large fine-tuned by RFTF achieved a score of $4.296$, surpassing the baseline of $4.283$, which also achieves new state-of-the-art performance. 

\subsubsection{Adaptation}
\label{Adaptation}
\begin{table}[t]
    \centering
    \caption{\textbf{Adaptation Experiments.} VLA refers to models to be fine-tuned. Env denotes the environments we use to fine-tune the model, where ``–'' indicates no fine-tuning.}
    \label{adaptation_results}
    \scalebox{1}{
        \begin{tabular}{c|c|ccccc|c}
            \toprule
            \multirow{2}{*}{VLA} & \multirow{2}{*}{Env} & \multicolumn{6}{c}{Task completed in a row}  \\
            \cmidrule{3-8}
            & & L1 & L2 & L3 & L4 & L5 & Avg. Len. $\uparrow$\\
            \midrule
            GR-MG & - & 96.8 & 89.3 & 81.5 & 72.7 & 64.4 & 4.047 \\
            GR-MG & ABC & 96.9 & 88.8 & 82.1 & 74.9 & 65.4 & 4.081 \\
            \textbf{GR-MG} & \textbf{D} & \textbf{96.1} & \textbf{90.5} & \textbf{83.9} & \textbf{75.0} & \textbf{65.8} & \textbf{4.113} \\
            \midrule
            Seer-Large & - & 96.3 & 91.6 & 86.1 & 80.3 & 74.0 & 4.283 \\
            Seer-Large & ABC & 96.4 & 91.7 & 86.7 & 80.7 & 74.1 & 4.296 \\
            \textbf{Seer-Large} & \textbf{D} & \textbf{97.0} & \textbf{92.0} & \textbf{86.0} & \textbf{80.6} & \textbf{74.5} & \textbf{4.301} \\
            \bottomrule
        \end{tabular}
    }
\end{table}

% In experiments focused on adapting embodied agents to unfamiliar environments, we fine-tune the embodied agents in test environments that they have never seen before. 
In this experiment, we demonstrate that embodied agents can be adapted to new environments that they have never seen before with the proposed RFTF.
Specifically, we fine-tune models with RFTF in the unseen D environment of CALVIN. As indicated in Table \ref{adaptation_results}, GR-MG fine-tuned by RFTF achieved a score of $4.113$, and Seer-Large fine-tuned by RFTF achieved a score of $4.301$, significantly outperforming the model's original performance in CALVIN's D environment. Since other methods are exclusively trained on the A, B, and C environments in CALVIN and have no exposure to the D environment, we refrain from comparing them with the results in our adaptation setting.

% Regarding implementation details, we first discretized the model’s output range into 1000 bins to obtain the probability term in the PPO optimization objective. To enhance training stability, we froze the model’s encoders and transformer backbone, fine-tuning only the action head. 
% During RL fine-tuning, we used four A40 GPUs, fine-tuning the model for approximately 10 hours with a learning rate of 1e-7, covering roughly 1000 episodes. 
% During RL fine-tuning, we train the model with a learning rate of 1e-7, covering roughly 1000 episodes. 
% The RL fine-tuning process is done with four NVIDIA A40 GPUs within 10 hours.
% To prevent overfitting to task instructions, we deliberately used different instructions during the fine-tuning phase compared to the testing phase.

% 这个图有点小，线可以粗一些，换深一点的颜色，或者可以横向扩大；把loss曲线也放进来 DONE
\subsection{Analysis of value model}
\label{Analysis of value model}
\begin{figure}[t]
    \centering
    \includegraphics[width=1\linewidth]{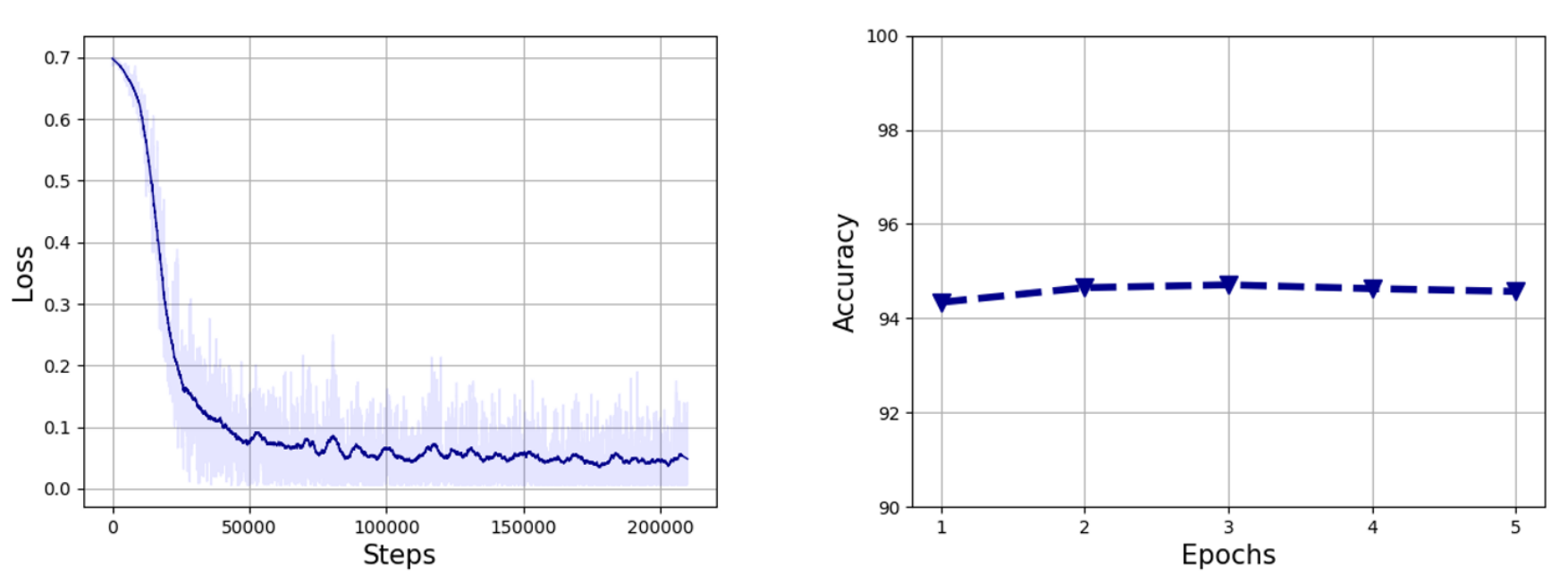}
    \caption{
    \textbf{Train curves of the value model.}
    We show the loss curve of the value model during training and evaluation results of the value model across different epochs.}
    \label{value_model_results}
\end{figure}
% We trained the value model on the CALVIN ABC training set with a batch size of 4$\times$8 and a learning rate of 1e-5. 
To verify whether the value model can accurately predict state values, we tested its accuracy on the CALVIN ABC validation set. Specifically, we selected two frames from expert-demonstrated trajectories, and if the value model assigned a higher value to the later frame compared to the earlier one, we considered the value prediction correct.

As shown in Figure \ref{value_model_results}, the value model achieved an accuracy of more than 94\% after the first epoch, with subsequent training nearly yielding no further improvements in accuracy. This observation aligns with findings in RLHF. To mitigate potential overfitting from prolonged training, we selected the value model from the first epoch for use in the reinforcement fine-tuning process.

\begin{figure}[t]
    \centering
    \includegraphics[width=1\linewidth]{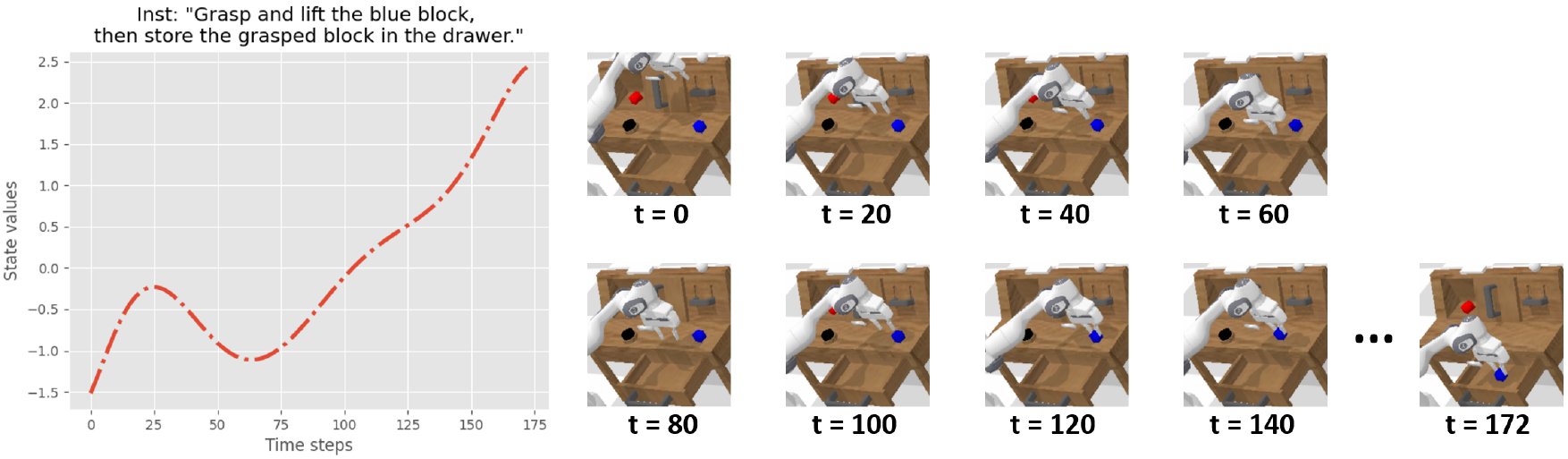}
    \caption{\textbf{An example of a state value curve.} As depicted, the curve exhibits a decline midway due to an incorrect grasping action by the embodied agent.}
    \label{episode_values}
\end{figure}

Notably, as shown in Figure \ref{episode_values}, we found that in episodes sampled by the embodied agent itself, the state values produced by the value model did not exhibit a monotonic increase over time. This ensures that the optimization objective of our reinforcement fine-tuning does not merely encourage the embodied agent to reinforce its original actions.

\subsection{Ablation study}
\label{Ablation study}
\begin{table}[t]
    \centering
    \caption{\textbf{Ablation studies.} VLA refers to models to be fine-tuned. SR refers to sparse reward. Replacing the dense rewards provided by RFTF with sparse rewards leads to performance drops.}
    \label{ablation_results}
    \renewcommand{\arraystretch}{1}
    \scalebox{1}{
        \begin{tabular}{c|c|c|ccccc|c}
            \toprule
            \multirow{2}{*}{VLA} & \multirow{2}{*}{Type} & \multirow{2}{*}{Reward} & \multicolumn{6}{c}{Task completed in a row}  \\
            \cmidrule{4-9}
            & & & 1 & 2 & 3 & 4 & 5 & Avg. Len. $\uparrow$\\
            \midrule
            GR-MG & Generalization & \textbf{RFTF} & \textbf{96.9} & \textbf{88.8} & \textbf{82.1} & \textbf{74.9} & \textbf{65.4} & \textbf{4.081} \\
            GR-MG & Generalization & SR & 95.3 & 88.3 & 80.8 & 72.1 & 62.8 & 3.993\\
            GR-MG & Adaptation & \textbf{RFTF} & \textbf{96.1} & \textbf{90.5} & \textbf{83.9} & \textbf{75.0} & \textbf{65.8} & \textbf{4.113} \\
            GR-MG & Adaptation & SR & 95.9 & 88.3 & 79.8 & 72.4 & 64.5 & 4.009\\
            \midrule
            Seer-Large & Generalization & \textbf{RFTF} & \textbf{96.4} & \textbf{91.7} & \textbf{86.7} & \textbf{80.7} & \textbf{74.1} & \textbf{4.296} \\
            Seer-Large & Generalization & SR & 95.2 & 89.9 & 85.1 & 79.4 & 72.9 & 4.225\\
            Seer-Large & Adaptation & \textbf{RFTF} & \textbf{97.0} & \textbf{92.0} & \textbf{86.0} & \textbf{80.6} & \textbf{74.5} & \textbf{4.301} \\
            Seer-Large & Adaptation & SR & 95.3 & 90.7 & 85.8 & 79.8 & 73.3 & 4.249\\
            \bottomrule
        \end{tabular}
    }
\end{table}

% table2有点乱，可以多加几列，把generalization/adaptation用type或者其他的表示，gr-mg和seer-large可以用vla表示，method改成reward？ DONE
% 这部分还可以说model generalization，说明我们的方法可以适配各种vla
We conducted ablation experiments to evaluate the effectiveness of the dense rewards in RFTF. 
To ensure a fair and controlled comparison, all experimental conditions were kept identical except for the rewards used for fine-tuning. 
%
% Specifically, the control group was trained using only sparse rewards that were derived solely from whether the agent successfully completed the given task or not.
The sparse rewards for fine-tuning are derived solely from whether the agent successfully completed the given task or not.

As shown in Table \ref{ablation_results}, unlike models fine-tuned with RFTF, the models using only sparse rewards exhibited varying degrees of performance drop. This finding is consistent with the observations in~\cite{hu2024flare} and~\cite{guo2025improving}.

\section{Conclusion and limitation}
\label{Conclusion and limitation}
In this paper, we propose RFTF, an online reinforcement fine-tuning method for embodied agents. 
To obtain dense rewards, we first train a value model using temporal information while maintaining low data dependency. 
Then, we integrate the value model into the reinforcement fine-tuning process for embodied agents, providing reward signals for intermediate decision steps, addressing the prevalent issue of sparse rewards, and enhancing the effectiveness of fine-tuning. 
Experimental results demonstrate that embodied agents fine-tuned with RFTF exhibit superior generalization and overall performance. Additionally, RFTF enables rapid adaptation to new environments. 
% The primary limitation of RFTF is the necessity to freeze the encoders and transformer backbone during fine-tuning, and we currently lack the hardware resources to support real-world experiments.
The primary limitation of RFTF is that it has only been verified on the simulated benchmark. In the future, we will apply RFTF to real-world robots.

\bibliographystyle{plain}
\bibliography{reference}

\end{document}